\documentclass[review]{elsarticle}

\usepackage{ragged2e}
\usepackage{booktabs}
\usepackage[table,xcdraw]{xcolor}
\usepackage{graphicx}
\usepackage{array}
\usepackage{amssymb}
\usepackage{amsthm}
\usepackage{algorithm}
\usepackage{algpseudocode}
\usepackage{amsmath}
\usepackage{multirow}
\usepackage{geometry}
\usepackage{threeparttable}
\usepackage{algorithm,algpseudocode,float}
\usepackage{lipsum}

\UseRawInputEncoding

\usepackage{lineno,hyperref}

\makeatletter
\newenvironment{breakablealgorithm}
{
		\begin{center}
			\refstepcounter{algorithm}
			\hrule height.8pt depth0pt \kern2pt
			\renewcommand{\caption}[2][\relax]{
				{\raggedright\textbf{\ALG@name~\thealgorithm} ##2\par}%
				\ifx\relax##1\relax 
				\addcontentsline{loa}{algorithm}{\protect\numberline{\thealgorithm}##2}%
				\else 
				\addcontentsline{loa}{algorithm}{\protect\numberline{\thealgorithm}##1}%
				\fi
				\kern2pt\hrule\kern2pt
			}
		}{
		\kern2pt\hrule\relax
	\end{center}
}
\makeatother

\begin{document}
	
\begin{frontmatter}
			
\title{RL-MSA: a Reinforcement Learning-based Multi-line bus Scheduling Approach}
		
\author[mymainaddress,mysecondaryaddress]{Yingzhuo Liu}

\author[mymainaddress,mysecondaryaddress]{Xingquan Zuo\corref{mycorrespondingauthor}}
\cortext[mycorrespondingauthor]{Corresponding author}
\ead{zuoxq@bupt.edu.cn}


\address[mymainaddress]{School of Computer Science, Beijing University of Posts and Telecommunications, Beijing, China.}
\address[mysecondaryaddress]{Key Laboratory of Trustworthy Distributed Computing and Service, Ministry of Education, China.}

\begin{abstract}
\emph{Multiple Line Bus Scheduling Problem} (MLBSP) is vital to save operational cost of bus company and guarantee service quality for passengers. Existing approaches typically generate a bus scheduling scheme in an offline manner and then schedule buses according to the scheme. In practice, \emph{uncertain events} such as traffic congestion occur frequently, which may make the pre-determined \emph{bus scheduling scheme} infeasible. In this paper, MLBSP is modeled as a Markov Decision Process (MDP). A \emph{Reinforcement Learning-based Multi-line bus Scheduling Approach} (RL-MSA) is proposed for bus scheduling at both the offline and online phases. At the offline phase, deadhead decision is integrated into bus selection decision for the first time to simplify the learning problem. At the online phase, deadhead decision is made through a time window mechanism based on the policy learned at the offline phase. We develop several new and useful state features including the features for control points, bus lines and buses. A \emph{bus priority screening mechanism} is invented to construct bus-related features. Considering the interests of both the bus company and passengers, a reward function combining the final reward and the step-wise reward is devised. Experiments at the offline phase demonstrate that the number of buses used of RL-MSA is decreased compared with offline optimization approaches. At the online phase, RL-MSA can cover all departure times in a timetable (i.e., service quality) without increasing the number of buses used (i.e., operational cost).
\end{abstract}
		
		
\end{frontmatter}
	
	
\section{Introduction}

\emph{Bus Scheduling Problem} (BSP) is a process of optimizing the allocation of buses to perform planned trips according to pre-determined bus timetables to achieve minimum operational cost while guaranteeing service quality \cite{2001applying}. The \emph{Multiple Line Bus Scheduling Problem} (MLBSP) is an important and challenging type of BSP that features multiple inter-dependent bus lines. Each bus line has two \emph{Control Points} (CPs). Every CP marks the departure location of a bus line. In MLBSP, a bus is required to start its trip from one CP of a bus line at a departure time indicated in the corresponding bus timetable of the CP. Meanwhile, in order to improve bus utilization and thereby reduce operational cost, a bus is sometimes required to travel from a CP of one bus line to another CP of a different bus line. This operation is called \emph{deadhead} in this paper \cite{2017dynamic}.

Any solution of a BSP can be obtained at two phases, i.e., the offline phase and the online phase. Existing approaches mainly focused on the offline phase and can be largely divided into two categories, exact approaches \cite{2001models, 1994column, 2006time} and heuristics \cite{Moreno2019, otsuki2016new, 2015clonal}. These approaches aim to generate complete \emph{bus scheduling schemes} offline based on the prior knowledge of all buses that can be scheduled as well as the bus timetables of every CP of each bus line. However, whenever \emph{uncertain events} such as traffic congestion occurs, such pre-determined bus scheduling schemes may fail since buses can arrive at any CPs with significant delays, affecting their next scheduled trips, resulting in poor and unresponsive performances.

To tackle this challenge, recently a few studies have started to solve BSPs at the online phase \cite{2020multi}. Existing online bus scheduling approaches can be largely divided into two categories, robust scheduling approaches \cite{2011stochastic, 2016probabilistic, 2012robust} and rescheduling approaches \cite{2020multi, 2004robust, 2019robust}. Robust scheduling approaches aim to generate robust bus scheduling schemes that can maintain their feasibility despite of delayed trips by overestimating the travel time of some scheduled trips \cite{2011stochastic}. However, a robust bus scheduling scheme will inevitably use more buses than necessary, resulting in increased operational cost. Therefore, to improve bus utilization, rescheduling approaches adopt a different strategy of regenerating the bus schedule whenever unexpected events occur. However, uncertain events that can disrupt existing scheduling schemes may occur frequently in practice , causing frequent rescheduling and increased computation cost. While facing challenges, bus scheduling at the online phase is undoubtedly important for a bus company to reduce its operation cost while ensuring the service quality \cite{2001applying}.

\emph{Reinforcement Learning} (RL) is a booming machine learning technology that has been widely utilized to solve many complex sequential decision-making problems \cite{huang2020deep, samir2021optimizing, liu2023reinforcement}. Compared with traditional optimization approaches, an RL agent is designed to learn a policy through a trial-and-error process by interacting directly with its learning environment. The trained policy is then exploited by the agent to solve any decision-making problems with the maximum possible performance. Among all machine learning paradigms, RL has the unique capability of handling uncertainty events and therefore is well-suited for MLBSP as a typical uncertain environment. 

Motivated by the above, RL technologies will be extensively explored in this paper to tackle the MLBSP. Specifically, we propose a novel \emph{Reinforcement Learning based Multi-line bus Scheduling Approach} (RL-MSA) for bus scheduling at both the offline and online phases. 

As far as we are aware, RL-MSA is different from all existing methods for BSP.  Different from existing approaches which generate a complete scheduling scheme directly, in RL-MSA, each departure time indicated in a bus timetable is treated as a decision point. An RL agent can make two types of decisions at any decision point. They are (1) the bus selection decision that chooses a bus to perform the scheduled trip, and (2) the deadhead decision that asks a bus to travel from its current CP to a CP of a different bus line. The MLBSP is subsequently modeled as a \emph{Markov Decision process} (MDP) \cite{1957markovian}. To address this MDP problem, our RL agent is designed to use the state-of-the-art \emph{Proximal Policy Optimization} (PPO) algorithm for policy training. 

To enhance the effectiveness of PPO, we develop new state features in this paper, including the features for buses, CPs and bus lines. Meanwhile, a new \emph{bus priority screening mechanism} is invented to construct bus-related features, effectively reducing the dimensionality of the state space. Furthermore, to boost learning speed, we design a new reward function to provide informative and immediate feedback to the learning agent at each decision point by combining the final reward with a new step-wise reward.

We further create appropriate processes respectively for bus scheduling at the offline and online phases. Particularly, at the offline phase, to simplify the learning problem, deadhead decision is integrated into the bus selection decision. On the other hand, at the online phase, two types of decisions must be made separately. For effective learning, we propose to make the deadhead decision at the online phase by directly using the policy learned at the offline phase. This is achievable with the help of a new \emph{time window mechanism}. Meanwhile, a separate policy is learned by the online RL agent to support smart bus selection decisions. In summary, through developing RL-MSA for both offline and online bus scheduling, this paper makes the following key contributions:

1) We propose the first MDP-based model of the MLBSP in the literature, laying the theoretical foundation for developing the RL-MSA in this paper. Different from most of existing methods for BSP that generate bus schedules offline, RL-MSA can effectively handle unexpected and disruptive events by dynamically adjusting the bus schedules online.

2) To achieve scalable and effective learning performance for RL-MSA, we design new state features specifically tailored for MLBSP, including the features of buses, CPs and bus lines, where the bus features are constructed based on a new bus priority screening mechanism. We also propose a new reward function that combines the final reward and the step-wise reward to expedite the process for our RL agent to learn useful bus scheduling policies. 
 
3) We create appropriate processes respectively for bus scheduling at the offline and online phases. To simplify the learning problem, deadhead decisions are integrated into bus selection decisions at the offline phase. Deadhead decision is further achieved at the online phase with the help of a new time window mechanism that allows us to directly use the policy trained at the offline phase
	
4) RL-MSA is compared experimentally with the state-of-the-art Adaptive Large Neighborhood Search (ALNS) method on 2 real-word problem instances and 8 artificial problems instances. Our experiments clearly show that RL-MSA can significantly outperform ALNS in terms of the total number of buses scheduled and the total deadhead time.

The remainder of this paper is organized as follows: Section 2 reviews the related work. Section 3 formally defines the MLBSP. In Section 4, the new RL-MSA is presented. Section 5 reports and analyzes the experimental results. Finally, conclusions are drawn in Section 6.

\section{Related works}

Bus scheduling problem is a complex optimization problem. Existing solution approaches can be divided into two categories: scheduling approaches for offline phase and scheduling approaches for online phase.

\subsection{Bus scheduling approaches for offline phase}
It is very time-consuming to generate bus scheduling schemes manually, and it also depends heavily on the scheduling experience of scheduler. Therefore, many scheduling approaches under static environment have been proposed. These approaches are mainly divided into two categories: exact approaches and heuristics.

(1) Exact approaches

There are many studies using exact approaches to solve BSP in offline phase. Exact approaches can obtain the optimal solution to a problem in an acceptable time when the problem size is small.  

In order to minimize operational cost, Freling et al. \cite{2001models} proposed a linear program combined with an auction-based-algorithm for BSP. Riberiro et al.\cite{1994column} modeled the BSP as a network flow model and proposed an approach based on a column generation algorithm to solve it. To minimize the operational cost, Kliewer et al. \cite{2006time} used a time-space network to model the BSP. Similarly, to reduce operational cost and bus exhaust emissions, Li et al. \cite{2009sustainability} modeled the BSP as a time-space network. Compared with the traditional network flow model, the time-space network model greatly reduces the model size, which can be solved using the CPLEX. 

Desaulnier et al. \cite{DESAULNIERS1998479} modeled the BSP with time windows as an integer nonlinear multi-commodity network flow model with time variables, and solved it using a column generation approach embedded in a branch-and-bound framework. For a BSP with time windows, Hadjar et al. \cite{2009dynamic} proposed a branch-and-bound algorithm solve it and a time windows reduction approach to speed up the search process. To reduce the fixed and dynamic cost, Haase et al. \cite{2001simultaneous} used the set partitioning formulation to model the simultaneous bus and crew scheduling problem, and solved it using a column generation process integrated into a branch-and-bound. To minimize the numbers of buses and drivers, Lin et al. \cite{2010bi} proposed a programming model solved by a branch-and-bound algorithm. 

Adler et al. \cite{2017vehicle} formulated the alternative-fuel BSP as an integer program, and proposed a branch-and-price algorithm to solve it. Gkiotsalitis et al. \cite{2022exact} modeled the electric BSP with time windows as a mixed-integer nonlinear programming model and then linearized the formulation of the problem by reformulating it to a mixed-integer linear program that can be solved optimally. Janovec et al. \cite{2019exact} proposed a linear programming model for the electric BSP and used a standard integer programming solver to solve it.

(2) Heuristics

Many heuristics have been proposed to solve BSP in offline phase. Heuristics obtain a suboptimal solution to the problem or optimal solution with a certain probability.

Moreno et al. \cite{Moreno2019} proposed a genetic algorithm for the BSP, and refined the solution by means of a set partitioning model. Otsuki et al. \cite{otsuki2016new} proposed a local search algorithm for BSP which utilizes pruning and deepening techniques in the variable depth search framework. Shui et al. \cite{2015clonal} combined a clonal selection algorithm with the DTAP to solve the BSP. Dell'Amico et al. \cite{dell1993heuristic} proposed a new polynomial-time heuristic algorithm using some structural properties of the BSP. To minimize the number of buses used, Ceder et al. \cite{2011public} proposed a deficit-function-based heuristic to solve the BSP. Kulkarni et al. \cite{2018new} proposed a column generation approach and three heuristics to solve a BSP. Pepin et al. \cite{pepin2009comparison} compared the performance of five different heuristics for BSP, namely, a truncated branch-and-cut method, a Lagrangian heuristic, a truncated column generation method, a large neighborhood search heuristic and a tabu search heuristic. Zhao et al. \cite{2022dp} proposed a tabu search algorithm for BSP.

To solve a multiple-type electric BSP, Yao et al. \cite{2020optimization} proposed a heuristic approach to generate a scheduling scheme to arrange recharging trips and departure time for each bus. Wang et al. \cite{wang2021solving} studied the electric BSP and proposed a genetic algorithm based column generation approach to solve it. For electric BSP, Wen et al. \cite{wen2016adaptive} proposed a mixed integer programming formulation as well as an adaptive large neighborhood search (ALNS) heuristic. Liu et al. \cite{liu2022two} proposed a two-stage solution approach combining a simulated annealing and a local search for electric BSP.

To simultaneously optimize bus timetabling and electric BSP, Teng et al. \cite{2020integrated} proposed a multi-objective particle swarm optimization algorithm to get a set of Pareto-optimal solutions. To simultaneously schedule buses and crews, Freling et al. \cite{2001applying} proposed an approach that combines a column generation algorithm and a Lagrangian relaxation algorithm. To minimize the number of drivers and buses, Zuo et al. \cite{2014vehicle} proposed an improved nondominated sorting genetic algorithm II (NSGA-II) combined with a departure times adjustment procedure (DTAP). Carosi et al. \cite{2019matheuristic} proposed a multi-commodity flow type model and a diving-type matheuristic approach for bus timetable optimization and BSP.  Desfontaines et al. \cite{desfontaines2018multiple} proposed a two-phase matheuristic: the first phase computes bus schedules with a column-generation heuristic; the second relies on a mixed integer program to find the best possible timetable considering the computed bus schedules.

\subsection{Bus scheduling approaches for online phase}
	
Uncertain events, such as traffic congestion and bus breakdown, may happen during the implementation of the scheduling scheme, resulting in the scheduling scheme becoming infeasible. Solution approaches for uncertain BSP can be divided into rescheduling approaches and robust scheduling ones.  

(1) Rescheduling approaches

Rescheduling approaches regenerate a new scheduling scheme to replace the old one. 

For a BSP under uncertain environment, Wang et al. \cite{2020multi} used NSGA-II to regenerate a scheduling scheme to deal with traffic congestion. For a BSP with uncertain travel time, Huisman et al. \cite{2004robust} proposed a cluster rescheduling heuristic. For a BSP under uncertain environment, Tang et al. \cite{2019robust} proposed a static model with buffer-distance strategy to tackle adverse impacts caused by trip time stochasticity, and used a dynamic model to periodically reschedule an electric bus fleet during a day’s operation. Shen et al. \cite{2017dynamic} proposed a scheduling approach based on a hierarchical task network and designed two dynamic bus scheduling strategies. The first one reschedules individual bus independently, and the second one reschedules multiple buses simultaneously. A robust rescheduling and holding approach \cite{2020robust} and a sequential hill climbing approach \cite{2021bus} were proposed to reschedule departure times of the trips to cope with uncertain events. Li et al. \cite{2008parallel} developed a sequential and parallel auction algorithm to solve an uncertain BSP. To minimize operation and delay cost of uncertain events, Li et al. \cite{2007decision} proposed a prototype decision support system that recommended solutions for a rescheduling BSP. 

(2) Robust scheduling approaches

Robust scheduling approaches generate a robust bus scheduling scheme, and improve the robustness of the scheduling scheme by overestimating the travel time of trips. 

To optimize planned cost and cost caused by disruptions, Naumann et al. \cite{2011stochastic} modeled the BSP as a time-space network, and presented a stochastic programming approach for robust BSP. To enhance the schedule's robustness, Shen et al. \cite{2016probabilistic} redefined the compatibility of any pair of trips and proposed a network flow model with stochastic trip times. To minimize the sum of the expected value of the random schedule deviation, Yan et al. \cite{2012robust} established a robust optimization model for a BSP with uncertain travel time, and a Monte Carlo simulation was used to solve the model.  

\subsection{Conclusion}

Overall, approaches for BSP under static environment generate a complete scheduling scheme offline, which may fail to ensure the scheduling quality and even become infeasible in the presence of uncertain events such as traffic congestion. In order to solve the uncertain events, the rescheduling approaches regenerate a scheduling scheme when uncertain events occur. However, when uncertain events occur frequently, frequent regeneration of the scheduling scheme will increase the operational cost. Robust scheduling approaches generate a robust scheduling scheme by overestimating the travel time of trips, which leads to lower bus utilization and increases the operational cost. Different from the existing approaches, this paper for the first time models the BSP as a MDP. Each departure time in the scheduling scheme is regarded as a decision point, and the RL is used to make decisions based on the information observed at each decision point.

\section{Bus scheduling problem}
	
The MLBSP studied in this paper focuses on multiple bus lines, and each line has two CPs. CP is the departure location of a line and also used for bus parking where there are a certain number of buses. Each CP has a timetable that includes a large number of departure times, and each departure time corresponds to a trip. A trip corresponds to the path of a bus line that a bus follows from one CP to another. Suppose the problem has $ n $ CPs, and the timetable of $ k $th CP is $ T_{k} = \left \{ t_{1}^{k}, t_{2}^{k},t_{3}^{k},\dots ,t_{N_{k} }^{k}\right \}  $, where $ N_{k} $ is the number of departure times in the $ k $th CP. The timetables of $ n $ CPs are combined into one timetable $ T $, and departure times in $ T $ are sorted in chronological order.
	
Suppose that the departure time of the $ i $th trip of a bus $ v $ is $ d_{i}^{v} $ and the arriving time is $ a_{i}^{v} $, then the travel time of the trip is $ h_{i}^{v} = d_{i}^{v} - a_{i}^{v} $. After completing a trip, the bus $ v $ can choose to perform a new trip, either for the current bus line or another bus line. When continuing performing trips of current line, the time between the arriving time of trip $ i $ and the departure time of the next trip ($ i +1$) is the rest time $ r_{i}^{v} = d_{i+1}^{v} - a_{i}^{v} $. Otherwise, it is necessary to perform a deadhead trip to the CP of another line and deadhead time is the time consumed by the deadhead trip. Assuming the deadhead time after the trip $ i $ is $ k_{i}^{v} $, if the deadhead trip is not required, $ k_{i}^{v} = 0 $. 
	
Figure 1 presents an example of a bus travelling. This instance has two bus lines (line AB, line CD) with two CPs for each bus line. The CPs of line AB are CP-A and CP-B, and the CPs of line CD are CP-C and CP-D. The horizontal axis of the figure represents the time, the vertical axis represents the position of the bus. Each node represents a CP, corresponding to the start or end of a trip. As shown in the figure, the bus starts its first trip from CP-A of line AB, and performs the trip according to the timetable of line AB. The arrow from node 1 to node 1' is a trip. The arrow from node 1' to node 2 is the rest time. The arrow from node 2’ to node 3 is a deadhead trip, the bus arrives at the CP-C of line CD and performs the trip according to the timetable of line CD.
	
The constraints of MLBSP are as follows:

1) Bus $ v $ must finish any ongoing trip $ i $ before it can start to take its next trip ($ i+1 $).
	
2) The rest time of the bus $ v $ after finishing the trip $ i $ cannot be less than the minimum rest time $ r_{min} $, i.e., $ r_{i}^{v} \ge  r_{min} $.
	
3) After the bus $ v $ completes the trip $ i $ of one bus line, if the next trip $ (i+1) $ is a trip of another bus line, it shall satisfy the condition: $ r_{min} + k_{i}^{v} \le  d_{i+1}^{v} - a_{i}^{v} $.
	
4) All departure times in the combined timetable $ T $ should be covered, i.e., one bus departs at each departure time in $ T $. 
	
The objectives of MLBSP are as follow:
	
1) Minimize the total number of buses used, ${N_{u}}$.
	
2) Minimize the total deadhead time $ T_{d} $, $ T_{d} = {\textstyle \sum_{v=1}^{N_{u}}}  {\textstyle \sum_{i=1}^{N^{v}}} k_{i}^{v}  $.

\begin{figure}[H]
	\centering
	\includegraphics[height=6cm,width=12cm]{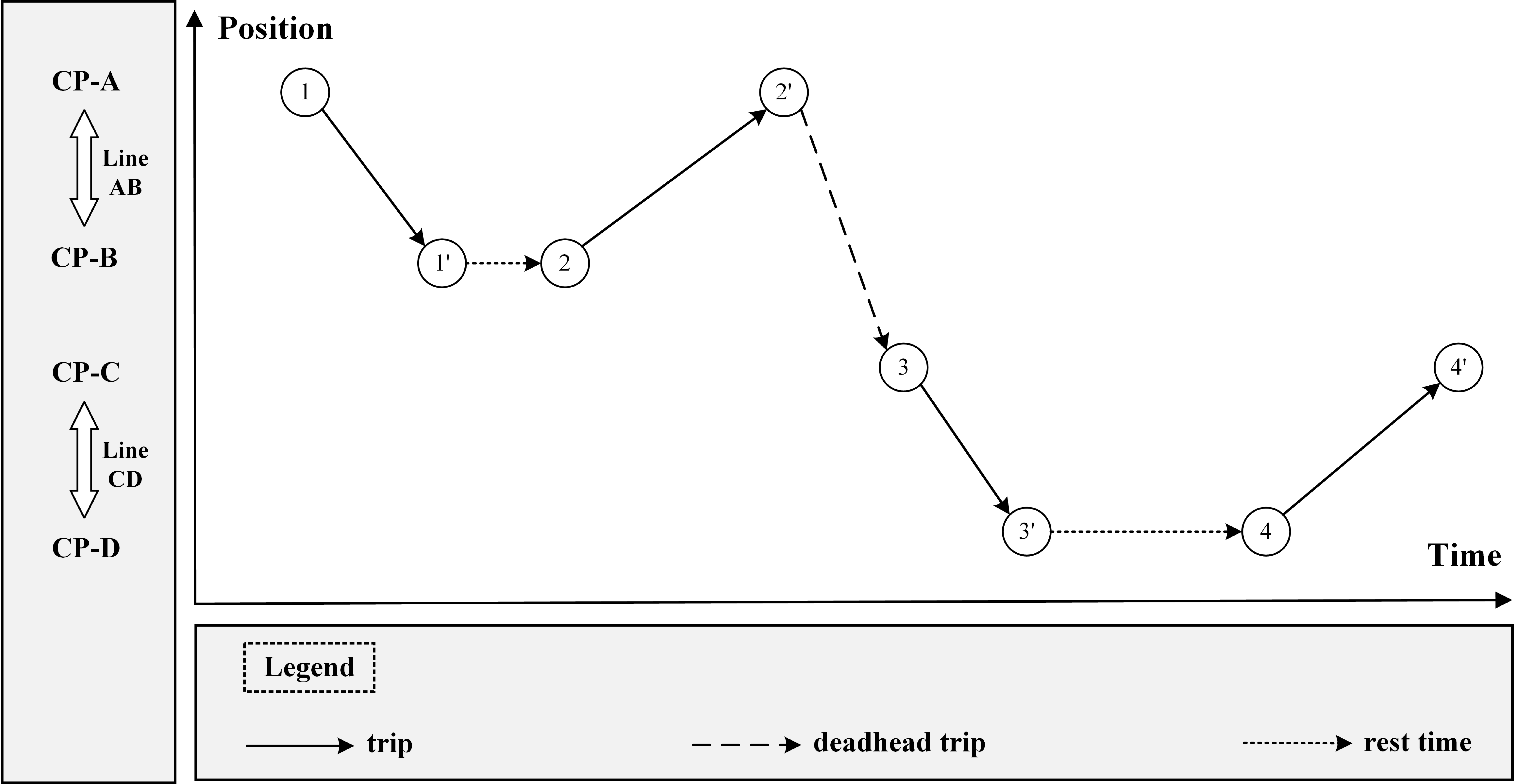}
	\centering
	\caption{An example of a bus travelling}
\end{figure}
	
\section{Reinforcement learning-based bus scheduling approach}
	
MLBSP is modeled as a MDP. With each departure time in the timetable $ T $ as a decision point, RL agent selects a bus to depart at the departure time according to the information of buses, CPs and bus lines. 

The framework of RL-MSA is shown in Figure 2. The environment can be supported by a simulator for MLBSP, which can update the information of buses, CPs and bus lines according to the decisions of the agent. The agent constructs the state features (\textbf{Subsection 4.1.1}) according to the information of buses, CPs and bus lines obtained from the environment, where the features of buses are constructed based on the bus priority screening mechanism. Bus priority screening mechanism is the process of setting priority for buses and screening buses according to the priority. The agent outputs the probability corresponding to all actions (\textbf{Subsection 4.1.2}) based on the state using the actor network (Actor-Net, which will be introduced in \textbf{Subsection 4.2.1}) of PPO. Then, the action is selected according to the probability. After the selected action is executed in the environment, a new state and the corresponding reward (\textbf{Subsection 4.1.3}) can be obtained from the environment. 
 
\begin{figure}[H]
	\centering
	\includegraphics[height=10cm,width=9cm]{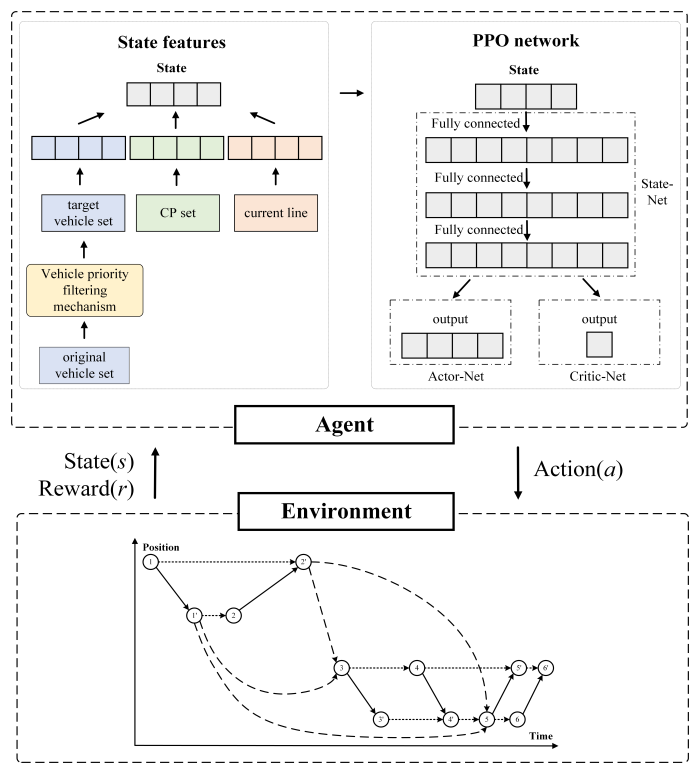}
	\centering
	\caption{Framework of RL-MSA}
\end{figure}

RL-MSA can be used at the offline phase and online phase. At the offline phase, a complete scheduling scheme is generated in advance using RL agent and buses are scheduled according to the scheme. At the online phase, RL agent makes online decisions at each departure time based on real-time information. 

As shown in Figure 3, at the offline phase, to avoid making bus selection and deadhead decisions at the same time, deadhead decision is integrated into the bus selection decision. Therefore, only one offline agent is trained to make bus selection decisions. At the online phase, two types of decisions must be made separately. Therefore, both bus selection and deadhead decisions need to be made by their own agent. Since bus selection and deadhead are two types of different decisions, and deadhead decision is quite complex, we do not use the multi-agent approaches to jointly train the two agents. When training the online agent for bus selection, the agent is only responsible for bus selection decision, and deadhead decision is made by the online agent for deadhead. What's more, online agent for deadhead makes deadhead decision based on the trained offline agent. This is achievable with the help of a new time window mechanism. Therefore, there is no need to model the problem as a MDP for the offline (online) agent for deadhead. The MDP model discussed below is for offline (online) agent for bus selection.

\begin{figure}[H]
	\centering
	\includegraphics[height=6.5cm,width=10cm]{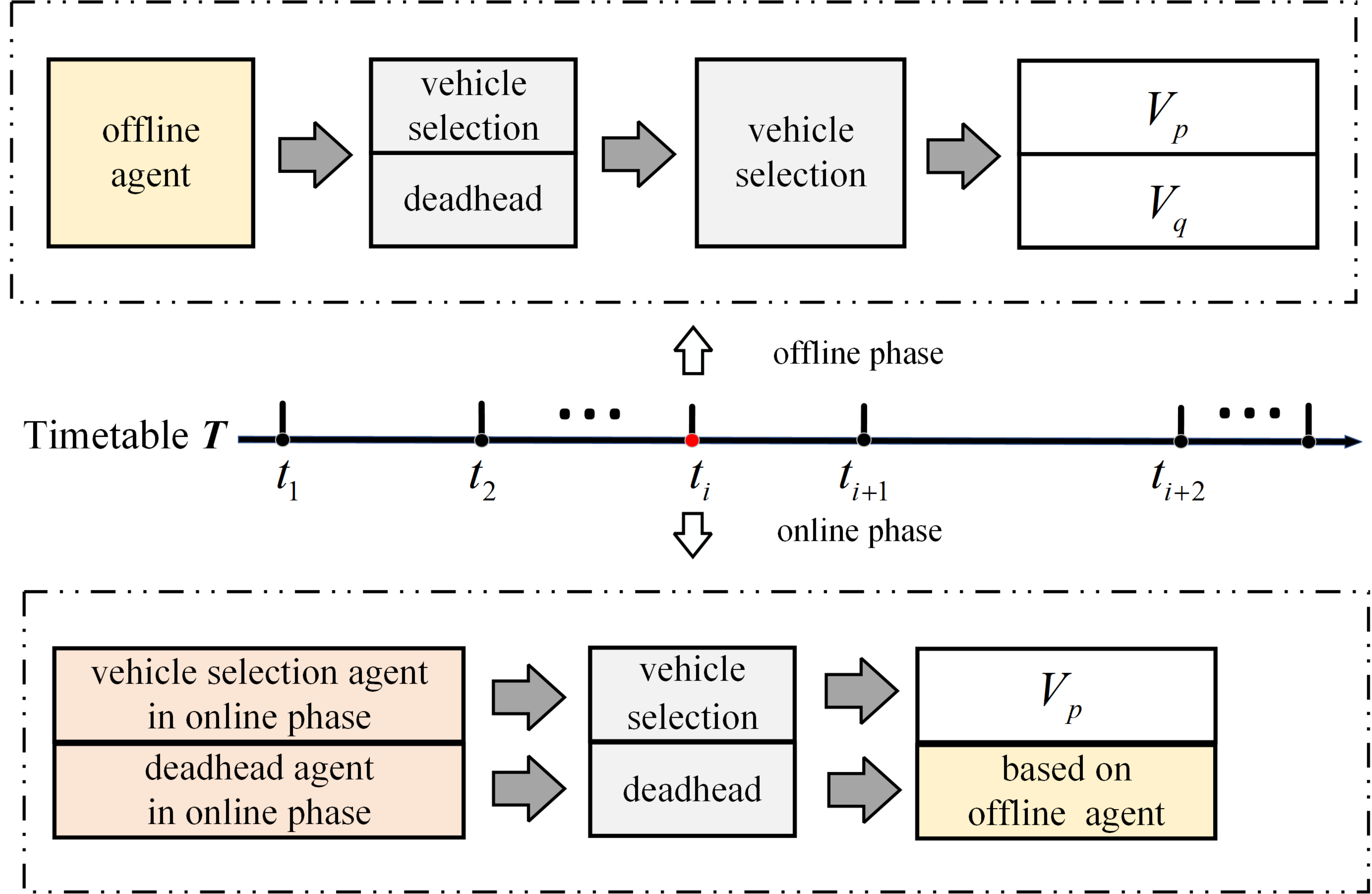}
	\centering
	\caption{Actions of MLBSP}
\end{figure}
 
\subsection{MDP model of RL-MSA}

In order to tackle MLBSP, the first step towards building RL-MSA is to model the problem as a MDP with clearly defined state space, action space and reward function

\subsubsection{State space}

The state devised in RL-MSA includes the features of buses, CPs and bus lines.   

(1) The features of CPs

Each bus line has two CPs, and the following features are selected for each CP $ c $. 
	
a. CP number, $ o^{c} $.

b. The total number of future trips of CP $ c $, $ n_{l}^{c} $: the number of departure times indicated in the corresponding timetable of the CP from the current moment to the last moment (each departure time corresponds to a trip). It reflects the future demand for the buses of the CP $ c $.

c. The number of short-term trips of CP $ c $, $ n_{s}^{c} $: the number of departure times indicated in the corresponding timetable of the CP for the next 300 minutes from the current moment. It reflects the future demand for buses of the CP $ c $ in the short term.

d. The number of buses at CP $ c $, $ n_{a}^{c} $: the total number of buses at CP $ c $ that can be selected to perform a new trip at a given departure time.

e. The number of buses that have performed at least a trip at CP $ c $, $ n_{o}^{c} $: $ n_{o}^{c} \le n_{a}^{c} $.

Feature a is used to obtain the correspondence between bus lines and CPs. Features b and c are used to obtain the future demand for buses of the CP. Features d and e are used to obtain the number of buses at the CP.

(2) The features of bus lines

Each departure time has a corresponding bus line $ l $, the following features are selected:

a. Departure CP of line $ l $, $ d^{l} $: the CP number of the departure CP of line $ l $.

b. Terminal CP of line $ l $, $ a^{l} $: the CP number of the terminal CP of line $ l $.

c. Travel time of line $ l $, $ h^{l} $: the travel time of the trip corresponding to the line $ l $.

Features a and b are used to obtain the correspondence between bus lines and CPs.

(3) The features of buses

For a bus $ v $, the following features are selected:

a. Whether bus $ v $ has been used, $ w_{s}^{v} $: $ w_{s}^{v}=1 $ means that the bus has performed at least a trip, which can be called a \emph{used bus}. Otherwise, the bus has not performed any trip, which can be called an \emph{unused bus}. In addition, the total number of buses $ N_{a} $ is limited, including both used and unused buses.

b. The rest time of bus $ v $, $ r^{v} $: the rest time of bus $ v $ after the last trip (deadhead trip). If $v$ is running, $ r^{v} = 0 $.

c. The CP where bus $ v $ currently locates, $ d^{v} $: the CP number of  the CP where $ v $ currently locates. If $v$ is running, $ d^{v} = -1 $.

d. The time required for the bus $ v $ to reach the departure CP of current departure time, $ k^{v} $: If the bus $ v $ is not at the departure CP of current departure time, the bus $v$ needs to perform a deadhead trip, and $ k^{v} $ is the deadhead time to arrive at the CP. If the bus does not need to perform a deadhead trip, $ k^{v}=0 $.

Feature a is used to obtain whether the bus is used or unused. Feature b is used to avoid long rest time and improve the bus utilization. Feature d is used to avoid buses from performing deadhead trips with large deadhead cost.

The specific state features are shown in Figure 4. The CP state contains the features of CPs for all bus lines. The bus line state contains the features of the bus line corresponding to current departure time. Buses are screened through the bus priority screening mechanism. The bus state contains the features of all buses in the bus set obtained by bus priority screening mechanism. This bus set is called \emph{target bus set} in this paper. Finally, the states of buses, CPs and bus lines are flattened and concatenated to obtain the final state.

The bus priority screening mechanism is different between the offline phase and online phase, which will be discussed separately. 

At the \textbf{offline phase}, in order to avoid making bus selection and deadhead decisions at the same time, deadhead decision is integrated into the bus selection decision. Four types of buses, the used (unused) buses that do not need to perform a deadhead trip and the used (unused) buses that need to perform a deadhead trip, are selected from all buses. Buses that have a rest time greater than the minimum rest time ($ r^{v}\ge r_{min} $), and locate at the departure CP of current departure time constitute the bus set $ V_{p} $, and buses in $ V_{p} $ are the buses that do not need to perform a deadhead trip. On the contrary, buses that do not locate at the departure CP of current departure time ($ k^{v} >0 $) and meet the Equation (1) constitute the bus set $ V_{q} $, and buses in $ V_{q} $ are the buses that need to perform a deadhead trip. Equation (1) is:

\begin{equation}\label{equ1}
	r^{v} > k^{v} + r_{min} 
\end{equation}
\\The rest time of the bus is greater than the sum of deadhead trip and the minimum rest time. As a result, these buses have sufficient free time to travel to the departure CP of current departure time. When a bus in $ V_{q} $ is selected, it first performs a deadhead trip to the departure CP of current departure time. Then, the bus can be selected to perform the trip corresponding to the current departure time once it arrives at the new CP. Therefore, when the bus selection decision is completed, the deadhead decision is also completed. What's more, both $ V_{p} $ and $ V_{q} $ contain used and unused buses.

After the buses are classified, internal sorting is carried out respectively for four types of buses. Since long rest time will lead to lower bus utilization, buses with long rest time need to be scheduled as soon as possible. Used buses that do not need to perform a deadhead trip (i.e., used buses in $ V_{p} $) are sorted according to the rest time. To reduce the cost of deadhead, the used buses that need to perform a deadhead trip (i.e., used buses in $ V_{q} $) are sorted according to the deadhead time. Note that, selecting any unused bus will have the same effect, it is not necessary for us to sort any unused buses. After completing the internal sorting of the same type of bus set, buses are selected from the corresponding bus set in the order of used buses that do not need to perform a deadhead trip, used buses that need to perform a deadhead trip, unused buses that do not need to perform a deadhead trip and unused buses that need to perform a deadhead trip, until the capacity of target bus set $ N_{s} $ is reached. 

At the \textbf{online phase}, at any departure time, the buses located at other CPs do not have sufficient time to travel to the departure CP of current departure time. This is different from the offline phase. Therefore, only buses in $ V_{p} $ can be selected, and deadhead decision cannot be integrated into the bus selection decision.

After the buses are classified, internal sorting is carried out respectively for used (unused) buses in $ V_{p} $. The rules of internal sorting are the same as the offline phase. After completing internal sorting, buses are selected from the corresponding bus set in the order of used buses that do not need to perform a deadhead trip and unused buses that do not need to perform a deadhead trip until $ N_{s} $ is reached. 

An example is shown in Figure 3. There are 9 buses in the bus set $ V_{p} $, including 4 used buses and 5 unused buses. There are 5 buses in the bus set $ V_{q} $, including 2 used buses and 3 unused buses. The capacity of target bus set $ N_{s} $ is 8. At the offline phase, 4 buses are selected from the used buses in $ V_{p} $ (bus 13, 4, 8, 26), 2 buses are selected from the used buses in $ V_{q} $ (bus 38, 2), and the remaining 2 buses are selected from the unused buses in $ V_{p} $ (bus 5, 14) finally. At the online phase, 4 buses are selected from the used buses in $ V_{p} $ (bus 13, 4, 8, 26), and the remaining 4 buses are selected from the unused buses in $ V_{p} $ (bus 5, 14, 27, 28).

\begin{figure}[H]
	\centering
	\includegraphics[height=12.1cm,width=12.4cm]{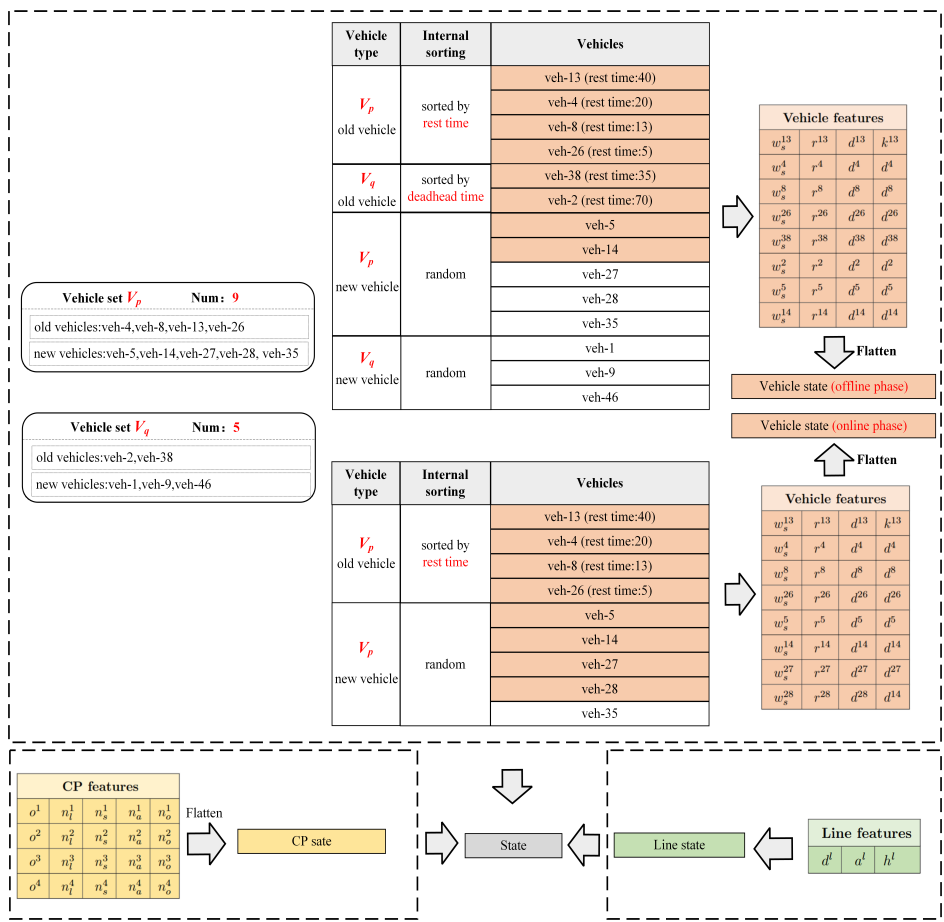}
	\centering
	\caption{State features of MLBSP}
\end{figure}

\subsubsection{Action space}
	
Action space differs at the offline phase and online phase, which will be discussed separately.

At the \textbf{offline phase}, deadhead decision is integrated into bus selection decision. As a result, only bus selection decision needs to be made. There are multiple buses which can be selected at each departure time, the RL agent needs to select a suitable bus among them, $ a\in A $. The action set $ A $ contains buses in $ V_{p} $ and $ V_{q} $. An example is shown in Figure 3, the target bus set with a capacity of 8 (bus 13, 4, 8, 26, 38, 2, 5, 14) is the action space at the offline phase. 

\textbf{Online phase} is different from the offline phase. At the online phase, as mentioned in Subsection 4.1.1, deadhead decision cannot be integrated into the bus selection decision. At each departure time, two types of decisions must be made independently.

In view of the above, the action space at the online phase is different from the action space at the offline phase. Specifically, we need to distinguish the action space for bus selection from the action space for deadhead. For the action space of bus selection, it only contains buses in $ V_{p} $. An example is shown in Figure 3, the target bus set with a capacity of 8 (bus 13, 4, 8, 26, 5, 14, 27, 28) is the action space for bus selection decision at the online phase.

For the deadhead decision, the RL agent must select several buses from the action space for deadhead at current departure time, and instruct these buses to travel to CPs of other bus lines. The action space for deadhead is the power set of the bus set consisting of buses with a rest time greater than the minimum rest time ($ r^{v}\ge r_{min} $) at all CPs at current departure time. To make the deadhead decision at the online phase, a time window mechanism is devised. A time window is fixed from the current departure time, and bus selection decisions are made at all departure times in the time window using the policy learned at the offline phase. Specifically, the following case exists: 
(1) A bus $ v $ that belongs to bus set $ V_{q} $ is selected at a departure time $ t_{j} $ in the time window.
(2) The bus needs to perform a deadhead trip to the departure CP of the departure time $ t_{j} $. (3) The \emph{estimated latest departure time} of the deadhead trip $ t_{e} $,

\begin{equation}\label{equ2}
	t_{e} = t_{j} -r_{min}-k^{v} 
\end{equation}
\\is between the current departure time $ t_{i} $ and the next departure time $ t_{i+1} $. 

In this case, the bus is dispatched at the current departure time $ t_{i} $ to perform a deadhead trip to arrive at the CP of the line corresponding to the departure time $ t_{i} $.

\subsubsection{Reward function}
	
Since the decision sequence of MLBSP is long, only using the final reward will lead to sparse reward problem, which makes it difficult for the agent to learn. Therefore, RL-MSA adopts the reward function combining the final reward and the step-wise reward. The final reward is set at the last step of the whole decision sequence, while the step-wise reward is set after each decision step.
	
(1) Final reward
	
The objective of MLBSP is to minimize the total number of buses used $ N_{u} $ and the total deadhead time $ T_{d} $. RL-MSA takes these two objectives into consideration to construct the main reward to evaluate the whole scheduling scheme.

The main reward is the weighted sum of the above two items:
\begin{equation}\label{equ3}
	r_{m} = -w_{1}^{1} \times N_{u} - w_{2}^{1}\times T_{d} 
\end{equation}	
\\where weight values $ w_{1}^{1} $, $ w_{2}^{1} $ are all positive real number.
	
(2) Step-wise reward

The step-wise reward contain many items, some of the items differ at the offline and online phase, and we will explain them separately. 

a. \textbf{offline phase}

To reduce the number of buses used, the agent is encouraged to avoid selecting unused buses when there are used buses eligible for the next trip. A penalty $ r_{n} $ is imposed if the agent select an unused bus.

To improve the utilization rate of buses, the agent is encouraged to select the buses with long rest time after the last trip (deadhead trip) $ r^{v} $, and a reward $ r_{k} $ is set. If the selected bus is an unused bus, $ r_{k} $ is set to 0. If the selected bus is a used bus, the number of used buses in $ V_{p} $ and $ V_{q} $, $ N_{o} $, is counted. These used buses are sorted in descending order according to the rest time $ r^{v} $. Assuming that the ranking of the selected bus $ v $ is $ p^{v} $, then $ r_{k} $ is defined as:

\begin{equation}\label{equ4}
	r_{k} = \frac{(N_{o} - p^{v} )}{N_{o} } 
\end{equation}	
\\The smaller $ p^{v} $ is, the longer the rest time of bus $ v $ is and the more it needs to be selected as soon as possible, so the reward $ r_{k} $ obtained by selecting this bus is larger.

To reduce the deadhead cost, and encourage the agent to select the bus with low deadhead cost, the deadhead time $ k^{v} $ is set as a penalty. 
	
To avoid performing a deadhead trip from a CP with a high future demand for buses to a CP with a low future demand for buses, when a bus in $ V_{q} $ is selected, the demand degree for buses $ U_{c} $ are calculated for the CP where the bus is currently located ($ U_{1} $) and the terminal CP of the bus line corresponding to current departure time ($ U_{2} $). For CP $ c $, the equation of $ U_{c} $ is as follows ($ n_{o}^{c} $ is in the denominator and may be equal to 0, so we add 1 to it):

\begin{equation}\label{equ5}
	U_{c} = \frac{n_{s}^{c}}{n_{o}^{c}+1}  
\end{equation}	
\\The larger $ n_{s}^{c} $ is, the higher the future demand for buses at CP $c$ is. The smaller $ n_{o}^{c} $ is, the less buses CP $c$ has. Finally, the larger $ U_{c} $ is, the more buses CP $c$ needs. If $ U_{1} $ is greater than $ U_{2} $, a penalty $ r_{u} $ ($ r_{u}=1 $) is imposed. When a bus in $ V_{p} $ is selected, the penalty term is 0. 

b. \textbf{online phase}

The online phase uses the same penalty  $ r_{n} $ as the offline phase.

For penalties $ k^{v} $ and $ r_{u} $, since there are no buses that need to perform a deadhead trip in the action space of the bus selection at the online phase, the two penalty terms do not need to be set, and both of them are 0.

For reward $ r_{k} $, the number of used buses in $ V_{p} $ is counted as $ N_{o} $ at the online phase. 

The step-wise reward is the weighted sum of the above four items:
	
\begin{equation}\label{equ15}
	r_{b}=-w_{1}^{2} \times r_{n}-w_{2}^{2} \times k^{v}+w_{3}^{2} \times r_{k}-w_{4}^{2} \times r_{u}
\end{equation}
\\where weight values $ w_{1}^{2} $, $ w_{2}^{2} $, $ w_{3}^{2} $ and $ w_{4}^{2} $ are all positive real number.

\subsection{Deep RL agent of RL-MSA}
	
The training data for the RL agent comes from the interaction between the agent and the simulated environment. The RL agent further trains its policy using the collected training data. Our RL agent makes use of the powerful feature extraction ability of deep neural network to cope with complex state and action spaces. The RL algorithm adopted by RL-MSA is PPO. PPO is a policy-based RL approach. Compared with value-based RL approaches, policy-based approaches have stronger exploration ability. In policy-based RL approaches, PPO is one of the classical and robust approaches, which has a wide range of applications \cite{yu2022surprising, huang2020deep, samir2021optimizing}.
	
\subsubsection{Network structure}
	
PPO is an approach based on the Actor-Critic framework. Both Actor and Critic have corresponding networks. As shown in figure 2, the network of Actor (Actor-Net) outputs the selection probability of each action in the action space, the network of Critic (Critic-Net) outputs the estimate value of current state (state value, i.e., $V$ value) and the feature extraction network (State-Net) is used to extract state features. To reduce the complexity of the network, Actor-Net and Critic-Net share the same State-Net, and State-Net is a four-layer fully connected network. The input of State-Net is the state features, and the output is a 32-dimensional vector, which serves as the input of Actor-Net and Critic-Net, respectively. Actor-Net finally outputs a vector with the same size as the action space, representing the corresponding selection probability of each action in the action space. Critic-Net finally outputs a scalar, representing the estimate value of the current state. 
	
\subsubsection{RL algorithm}

The main problem addressed by PPO is how to use the existing data to maximize the improvement of the policy without causing the policy to become worse due to excessive updating. Its main idea is to ensure the effectiveness of policy updating by limiting the gap between the unused policy and the used one.

RL-MSA adopts the clip version of PPO, that uses the following objective function:

\begin{equation}\label{equ6}
	L(s,a,\theta _{k},\theta)= \min (\frac{\pi_{\theta }(a|s) }{\pi_{\theta_{k}}(a|s) }A^{\pi _{\theta _{k} } }(s,a), g(\epsilon ,A^{\pi _{\theta _{k} } } (s,a)) 
\end{equation}
\\where,

\begin{equation}\label{equ7}
	g(\epsilon,A) =
	\left\{\begin{matrix}
		(1+\epsilon )A & A\ge 0 \\ 
		(1-\epsilon )A & A< 0 
	\end{matrix}\right.
\end{equation}

When $ A $ value is greater than 0, it means that the value of the selected action is higher than other actions. Therefore, the probability corresponding to the current action should be increased. However, in order to ensure the stability by preventing the updated policy does not deviate too much from the policy before updating. An upper limit is set for the ratio of the probability corresponding to the selected action of the updated policy and the policy before updating, i.e., $ (1+\epsilon ) $. Similarly, when $ A $ value is less than 0, it means that the value of the selected action is lower than other actions. Therefore, the probability corresponding to the current action should be decreased. Here, a lower limit is set for the ratio of the probability corresponding to the selected action of the updated policy and the policy before updating, i.e., $ (1-\epsilon ) $.

The pseudo code of PPO is shown in Algorithm 1, and can be divided into two parts: data collection and training. For the data collection part, firstly, the dataset is constructed  based on the interaction between Actor-Net and environment. Then, according to the rewards in the collected data, $ \hat{R_{t}} $ corresponding to each data in the dataset is calculated to train the Critic-Net. In addition, according to Critic-Net, the advantage value of each data in the dataset is calculated to train the Actor-Net. For the training part, first of all, the objective function for training Actor-Net is constructed for each data in the dataset according to equation (7), and the Adam optimizer is used for optimization. Then, the loss function for training Critic-Net is constructed, which is also optimized by Adam optimizer.
	
\begin{breakablealgorithm}
	\centering
	\caption{Learning process of PPO in RL-MSA}
	\begin{algorithmic}[1]
		\State \textbf{Parameters}: initial policy parameters $ \theta _{0} $, initial value function parameters $ \phi_{0} $.
		
		\For {k=0,1,2,...}
			\State Collect set of trajectories $ D_{k} =\left \{ \tau _{i}  \right \} $ by running policy $ \pi _{k} =\pi (\theta _{k} ) $ in the environment.
			\State Compute rewards-to-go $ \hat{R_{t} } $:

                $\hat{R_{t}}  = {\textstyle \sum_{i=t}^{T}}\gamma^{i-t}r_{t}$  
   
			\State Compute advantage estimates $ A_{t} $  based on the current value function $ V_{\phi _{k} }$:

                $A_{t} = \hat{R_{t}} - V_{t}$
            
			\State Update the policy by maximizing the PPO-Clip objectives:
			
			$\theta _{k+1} = \arg \min _{\theta }  \frac{1}{\left | D_{k}T  \right | }  \sum_{\tau \in D_{k} } \sum_{t=0}^{T} \min (\frac{\pi_{\theta }(a|s) }{\pi_{\theta_{k}}(a|s) }A^{\pi _{\theta _{k} } }(s_{t} ,a_{t} ), g(\epsilon ,A^{\pi _{\theta _{k} } } (s_{t} ,a_{t} ))$
			
			typically via stochastic gradient ascent with Adam.
			
			\State Fit value function by regression on mean-squared error:
			 
			 $\phi _{k+1} = \arg \min_{\phi} \frac{1}{\left | D_{k}  \right |T }\sum_{\tau \in D_{k} } \sum_{t=0}^{T} (V_{\phi }(s_{t} )-\hat{R_{t}}   )^{2}   $
			
			typically via stochastic gradient descent with Adam.
		
			\EndFor
			
	\end{algorithmic}
\end{breakablealgorithm}

\section{Experimental results}
	
In order to validate the effectiveness of RL-MSA, we apply it to two real-world MLBSP instances (Real-1, Real-2) found in Qingdao city, China and eight artificial MLBSP instances. To ensure that RL-MSA can cope with a variety of different MLBSP instances, artificial MLBSP instances are generated by randomly deleting different departure times from the instances Real-1 and Real-2 respectively (Artificial-1, Artificial-3, Artificial-5, Artificial-7 are generated from Real-2, Artificial-2, Artificial-4, Artificial-6, Artificial-8 are generated from Real-1).

RL-MSA is coded in python and runs on a PC with Intel (R) Core (TM) i5-13600KF CPU, 3.50 GHz, 32.00 GB RAM and NVIDIA RTX 3060 GPU. The weight values of the reward function of RL-MSA are shown in Table 1. The hyperparameters of PPO in RL-MSA are shown in Table 2. Table 3 presents the number of outputs and activation function of each layer of the components (State-Net, Actor-Net, Critic-Net) in PPO network.

\begin{table}[H]
	\centering
	\caption{Weight values in the final and step-wise rewards}
	\begin{tabular}{lllllll}
		\toprule[1pt]
		Parameters   & $ w_{1}^{1} $  & $ w_{2}^{1} $  & $ w_{1}^{2} $  & $ w_{2}^{2} $  & $ w_{3}^{2} $  & $ w_{4}^{2} $   \\ \toprule[0.7pt]
		Values       & 4.0  & 0.1  & 4.0  & 0.1  & 2.0  & 1.0    \\ \toprule[1pt]
	\end{tabular}
\end{table}
	
\begin{table}[H]
	\centering
	\caption{Hyperparameters of PPO}
	\begin{tabular}{ll}
		\toprule[1pt]
		Hyperparameters                         & Values   \\ \toprule[0.7pt]
		Hyperparameters of Adam            & 1e-5     \\ 
		Hyperparameters of Clip $\epsilon$ & 0.1      \\
		Discount factor $\gamma$           & 0.99      \\
		Learning rate                      & 1e-5      \\
		\toprule[1pt]
	\end{tabular}
\end{table}

\begin{table}[H]
	\centering
	\caption{Settings of the PPO network in RL-MSA}
	\begin{threeparttable}
		\begin{tabular}{llll}
			\toprule[1pt]
			Network	    				& Layer   & Number of outputs  & Activation function  \\ \toprule[0.7pt]
			\multirow{3}{*}{State-Net}	
			& 1       & 128    & ReLu    \\
			& 2       & 64     & Linear       \\    
			& 3       & 32     & ReLu    \\ 
			\multirow{1}{*}{Actor-Net}	    
			& 1       & 8    & Linear    \\
			\multirow{1}{*}{Critic-Net} 	
			& 1       & 1    & Linear    \\
			\toprule[1pt]
		\end{tabular}
	\end{threeparttable}
\end{table}

\subsection{Offline optimization of RL-MSA}
	
\begin{table*}[htp]
	\centering
	\caption{RL-MSA is superior to ALNS interms of the total number of buses and total deadhead time}
	\begin{tabular}{llllllllll}
		\toprule[1pt]
		\multicolumn{1}{l}{{Instances}} & \multicolumn{1}{l}{{Approaches}}  & \multicolumn{1}{l}{{$ N_{u} $}}  & \multicolumn{1}{l}{{$ T_{d} $}}  &\multicolumn{1}{l}{{$ N_{d} $}}  & \multicolumn{1}{l}{{Instances}} & \multicolumn{1}{l}{{Approaches}}  & \multicolumn{1}{l}{{$ N_{u} $}}  
		&\multicolumn{1}{l}{{$ T_{d} $}}  &\multicolumn{1}{l}{{$ N_{d} $}}  \\ \toprule[0.7pt]
			
		{\color[HTML]{0D0D0D} Real-1} & {\color[HTML]{0D0D0D} ALNS} & {\color[HTML]{0D0D0D} 57} & {\color[HTML]{0D0D0D} 921} & {\color[HTML]{0D0D0D} 0} & {\color[HTML]{0D0D0D} Real-2} & {\color[HTML]{0D0D0D} ALNS} & {\color[HTML]{0D0D0D} 95} & {\color[HTML]{0D0D0D} 1456}& {\color[HTML]{0D0D0D} 0} \\	
				
		{\color[HTML]{0D0D0D} } & {\color[HTML]{0D0D0D} RL-MSA} & {\color[HTML]{0D0D0D} 46} & {\color[HTML]{0D0D0D} 0} & {\color[HTML]{0D0D0D} 0} & {\color[HTML]{0D0D0D} } & {\color[HTML]{0D0D0D} RL-MSA} & {\color[HTML]{0D0D0D} 80} & {\color[HTML]{0D0D0D} 0}& {\color[HTML]{0D0D0D} 0} \\	
			
		{\color[HTML]{0D0D0D} Artificial-1} & {\color[HTML]{0D0D0D} ALNS} & {\color[HTML]{0D0D0D} 51} & {\color[HTML]{0D0D0D} 1016} & {\color[HTML]{0D0D0D} 0} & {\color[HTML]{0D0D0D} Artificial-2} & {\color[HTML]{0D0D0D} ALNS} & {\color[HTML]{0D0D0D} 48} & {\color[HTML]{0D0D0D} 1557}& {\color[HTML]{0D0D0D} 0} \\
				
		{\color[HTML]{0D0D0D} } & {\color[HTML]{0D0D0D} RL-MSA} & {\color[HTML]{0D0D0D} 38} & {\color[HTML]{0D0D0D} 894} & {\color[HTML]{0D0D0D} 0} & {\color[HTML]{0D0D0D} } & {\color[HTML]{0D0D0D} RL-MSA} & {\color[HTML]{0D0D0D} 42} & {\color[HTML]{0D0D0D} 582}& {\color[HTML]{0D0D0D} 0} \\	
					
		{\color[HTML]{0D0D0D} Artificial-3} & {\color[HTML]{0D0D0D} ALNS} & {\color[HTML]{0D0D0D} 80} & {\color[HTML]{0D0D0D} 1679} & {\color[HTML]{0D0D0D} 0} & {\color[HTML]{0D0D0D} Artificial-4} & {\color[HTML]{0D0D0D} ALNS} & {\color[HTML]{0D0D0D} 50} & {\color[HTML]{0D0D0D} 1083}& {\color[HTML]{0D0D0D} 0} \\
			
		{\color[HTML]{0D0D0D} } & {\color[HTML]{0D0D0D} RL-MSA} & {\color[HTML]{0D0D0D} 69} & {\color[HTML]{0D0D0D} 298} & {\color[HTML]{0D0D0D} 0} & {\color[HTML]{0D0D0D} } & {\color[HTML]{0D0D0D} RL-MSA} & {\color[HTML]{0D0D0D} 45} & {\color[HTML]{0D0D0D} 471}& {\color[HTML]{0D0D0D} 0} \\	
			
		{\color[HTML]{0D0D0D} Artificial-5} & {\color[HTML]{0D0D0D} ALNS} & {\color[HTML]{0D0D0D} 88} & {\color[HTML]{0D0D0D} 1461} & {\color[HTML]{0D0D0D} 0} & {\color[HTML]{0D0D0D} Artificial-6} & {\color[HTML]{0D0D0D} ALNS} & {\color[HTML]{0D0D0D} 42} & {\color[HTML]{0D0D0D} 949}& {\color[HTML]{0D0D0D} 0} \\
			
		{\color[HTML]{0D0D0D} } & {\color[HTML]{0D0D0D} RL-MSA} & {\color[HTML]{0D0D0D} 75} & {\color[HTML]{0D0D0D} 306} & {\color[HTML]{0D0D0D} 0} & {\color[HTML]{0D0D0D} } & {\color[HTML]{0D0D0D} RL-MSA} & {\color[HTML]{0D0D0D} 34} & {\color[HTML]{0D0D0D} 579}& {\color[HTML]{0D0D0D} 0} \\
			
		{\color[HTML]{0D0D0D} Artificial-7} & {\color[HTML]{0D0D0D} ALNS} & {\color[HTML]{0D0D0D} 70} & {\color[HTML]{0D0D0D} 1772} & {\color[HTML]{0D0D0D} 0} & {\color[HTML]{0D0D0D} Artificial-8} & {\color[HTML]{0D0D0D} ALNS} & {\color[HTML]{0D0D0D} 30} & {\color[HTML]{0D0D0D} 829}& {\color[HTML]{0D0D0D} 0} \\
			
		{\color[HTML]{0D0D0D} } & {\color[HTML]{0D0D0D} RL-MSA} & {\color[HTML]{0D0D0D} 60} & {\color[HTML]{0D0D0D} 612} & {\color[HTML]{0D0D0D} 0} & {\color[HTML]{0D0D0D} } & {\color[HTML]{0D0D0D} RL-MSA} & {\color[HTML]{0D0D0D} 28} & {\color[HTML]{0D0D0D} 588}& {\color[HTML]{0D0D0D} 0} \\
		\toprule[1.0pt]
	\end{tabular}
\end{table*}
	
For each instance, after training for 1000 episodes, the trained RL-agent is further tested on MLBSP instances at the offline phase. The following objectives are calculated: 1) the total number of buses used $ N_{u} $. 2) the total deadhead time $ T_{d} $. 3) the number of departure times uncovered $ N_{d} $. According to the optimization objectives of MLBSP, the smaller $ N_{u} $ ,$ T_{d} $ and $ N_{d} $ are, the better the performance will be. 
	
ALNS is a State-Of-The-Art (SOTA) approach for solving electric BSP. The performance of RL-MSA and ALNS are compared, as shown in Table 4. The trained RL agent has the same result for each run, thus, RL-MSA runs only once. ALNS is a stochastic approach, and the results of its each run are different. Thus, ALNS runs 30 times independently, and we take the best one of them.
	
As shown in Table 4, for all bus lines, the scheduling scheme obtained by RL-MSA has less $ N_{u} $, $ T_{d} $. In addition, RL-MSA also achieves the optimal results in $ N_{d} $ ($ N_{d}=0 $). In conclusion, the RL-MSA has better performance compared with ALNS.
	
The scheduling scheme of Real-1 generated by RL-MSA is shown in Figure 5. Different colors represent different bus lines. The scheduling scheme totally uses 46 buses and can cover all 452 departure times, without missing any departure time. Therefore, we believe RL-MSA can reliably meet the requirement of MLBSP at the offline phase. 

\begin{figure}[H]
	\centering
	\includegraphics[height=13cm,width=13cm]{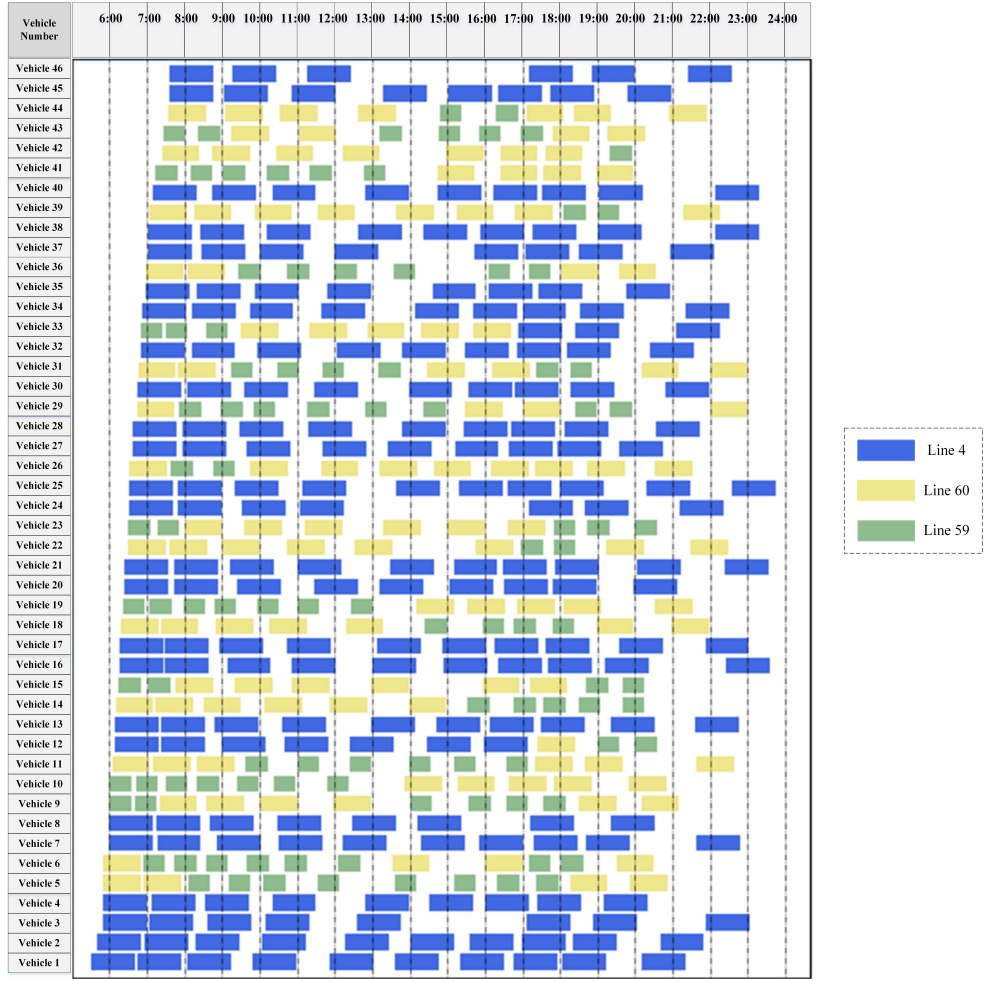}
	\centering
	\caption{A bus scheduling scheme generated by RL-MSA for Real-1}
\end{figure} 
 
\subsection{Online optimization of RL-MSA}

Uncertain events such as abnormal weather, traffic accidents and traffic congestion will prolong the travel time of some trips. For approaches for the offline phase, when the travel time becomes longer, the buses cannot return to CP as planned, so the buses cannot be dispatched according to the original scheduling scheme. In this case, the original scheduling scheme can only be adjusted by dispatching additional buses temporarily. RL-MSA can schedule buses in an online manner based on real-time information to effectively handle uncertain events. When no uncertain events occur, the buses are scheduled according to the scheme generated in advance. When uncertain events occur, the RL agent starts to schedule buses in an online manner. 

To validate that RL-MSA can cope with uncertain events, we conduct the following two experiments. In the first experiment, we add 15 minutes to the travel time of all trips between 13:30 pm and 16:40 pm. In the second experiment, we add 15 minutes to the travel time of all trips of line 59 during the whole day. The results of the two experiments are shown in Figure 6 and Figure 7, respectively. Figure 6 (a) and Figure 7 (a) are the original scheduling scheme of Real-1. Figure 6 (b) is the scheduling scheme adjusted by RL agent in experiment 1 and Figure 7 (b) is the scheduling scheme adjusted by RL agent in experiment 2. The performance of two scheduling schemes adjusted by RL agent is shown in Table 5. When traffic congestion occurs, the agent maintains the full coverage of all departure times in timetable by adjusting the departure order and departure time of some buses without increasing the number of buses used and the total deadhead time.	
	
\begin{table}[H]
	\centering
	\caption{the performance of two scheduling schemes adjusted by RL agent}
	\begin{tabular}{llll}
		\toprule[1pt]
		Scheme       & $ N_{u} $     & $ T_{d} $     & $ N_{d} $  \\ \toprule[0.7pt]   
		Original      & 46            & 0        & 0        \\
		6 (b)         & 46            & 0        & 0        \\
            7 (b)         & 46            & 0        & 0        \\
		\toprule[1pt]
	\end{tabular}
\end{table}		

\begin{figure}[H]
	\centering
	\includegraphics[height=19cm,width=12.3cm]{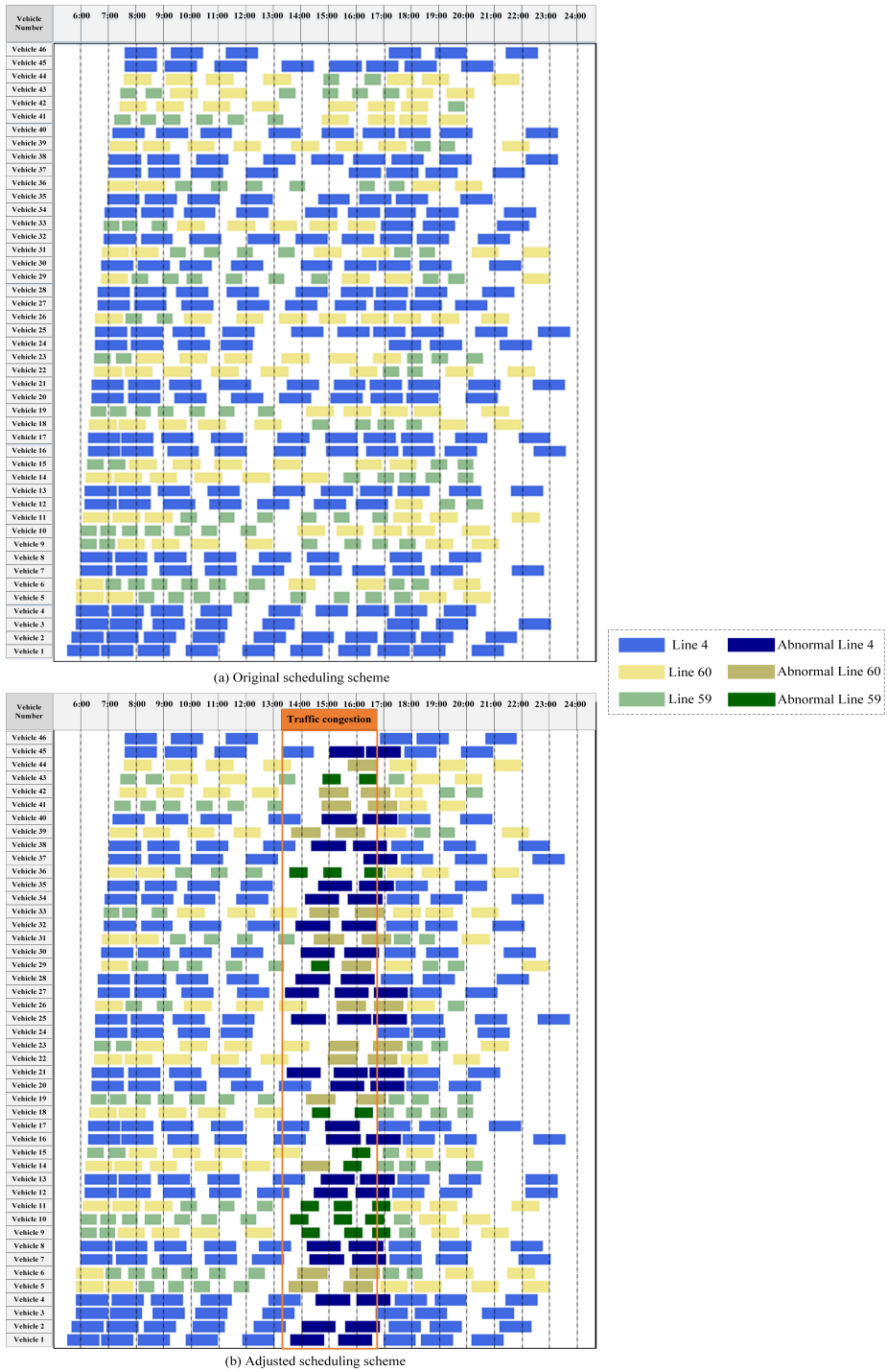}
	\centering
	\caption{experiment 1}
\end{figure}

\begin{figure}[H]
	\centering
	\includegraphics[height=19cm,width=11cm]{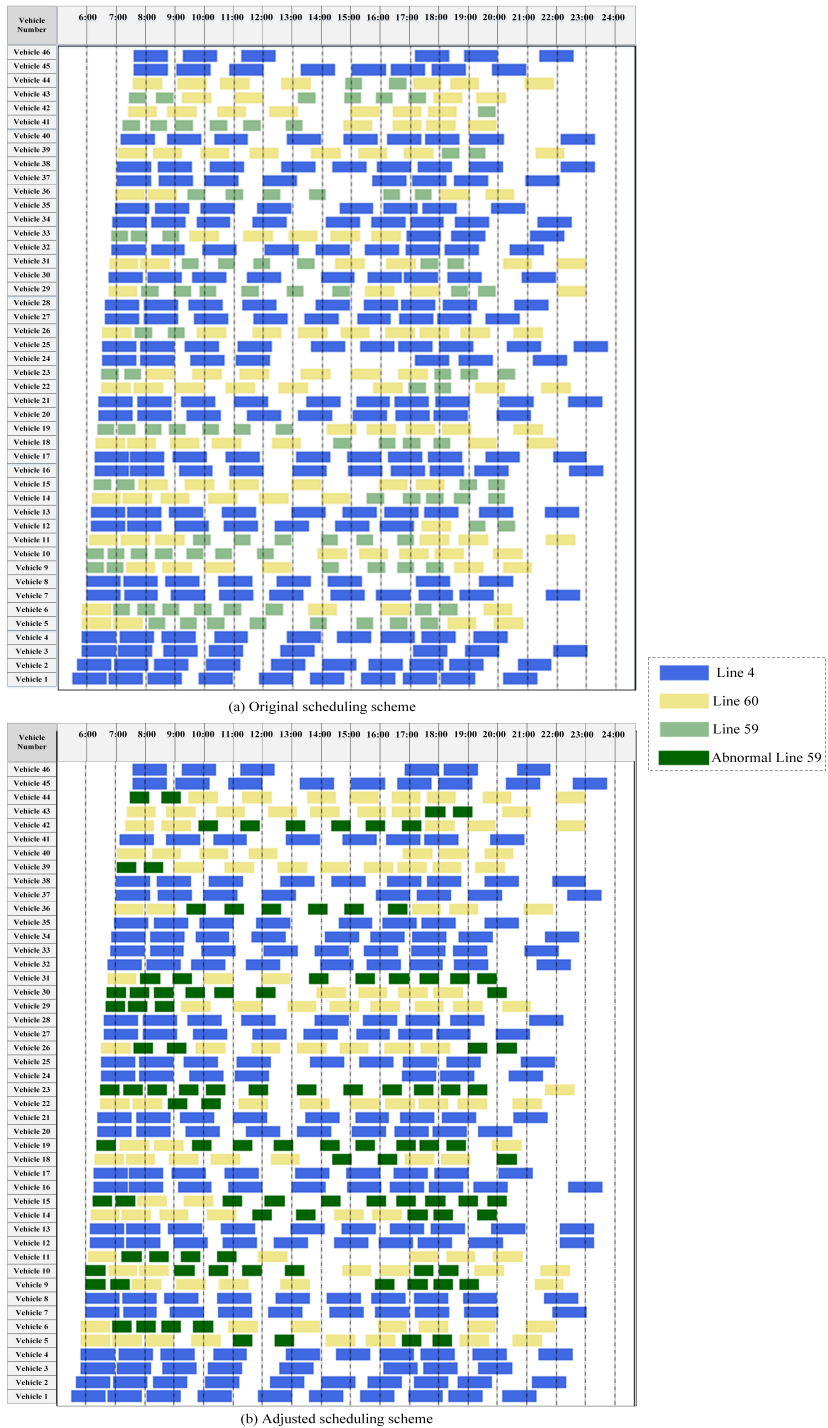}
	\centering
	\caption{experiment 2}
\end{figure}

\subsection{Ablation experiments}
	
\subsubsection{RL algorithm}
To explore the impact of RL algorithms on RL-MSA, four RL algorithms (DQN, D3QN, REINFORCE, PPO) are selected, RL-MSA-D, RL-MSA-E and RL-MSA-R are constructed by replacing PPO in RL-MSA with DQN, D3QN and REINFORCE, respectively. This paper conducts experiments on Real-1, and all approaches are trained for 1000 episodes. The convergence curves of the four approaches are shown in Figure 8 (horizontal axis is the episode, and vertical axis is the accumulated reward). The three objectives achieved by the four approaches are shown in Table 6. Figure 8 shows that all the 4 approaches finally converge. The convergence of the two approaches using policy-based algorithms (RL-MSA, RL-MSA-R) is faster than that of the two approaches using value-based algorithms (RL-MSA-D, RL-MSA-E). In addition, RL-MSA finally converges to a better result than RL-MSA-R. According to the table, RL-MSA and RL-MSA-D achieve the lowest total number of buses used. RL-MSA achieves the best result in the total deadhead time, and all the four approaches can cover all the departure times in the timetable. In conclusion, the proposed approach achieves the best results when using PPO.
	
\begin{figure}[H]
	\centering
	\includegraphics[height=8.2cm,width=11.3cm]{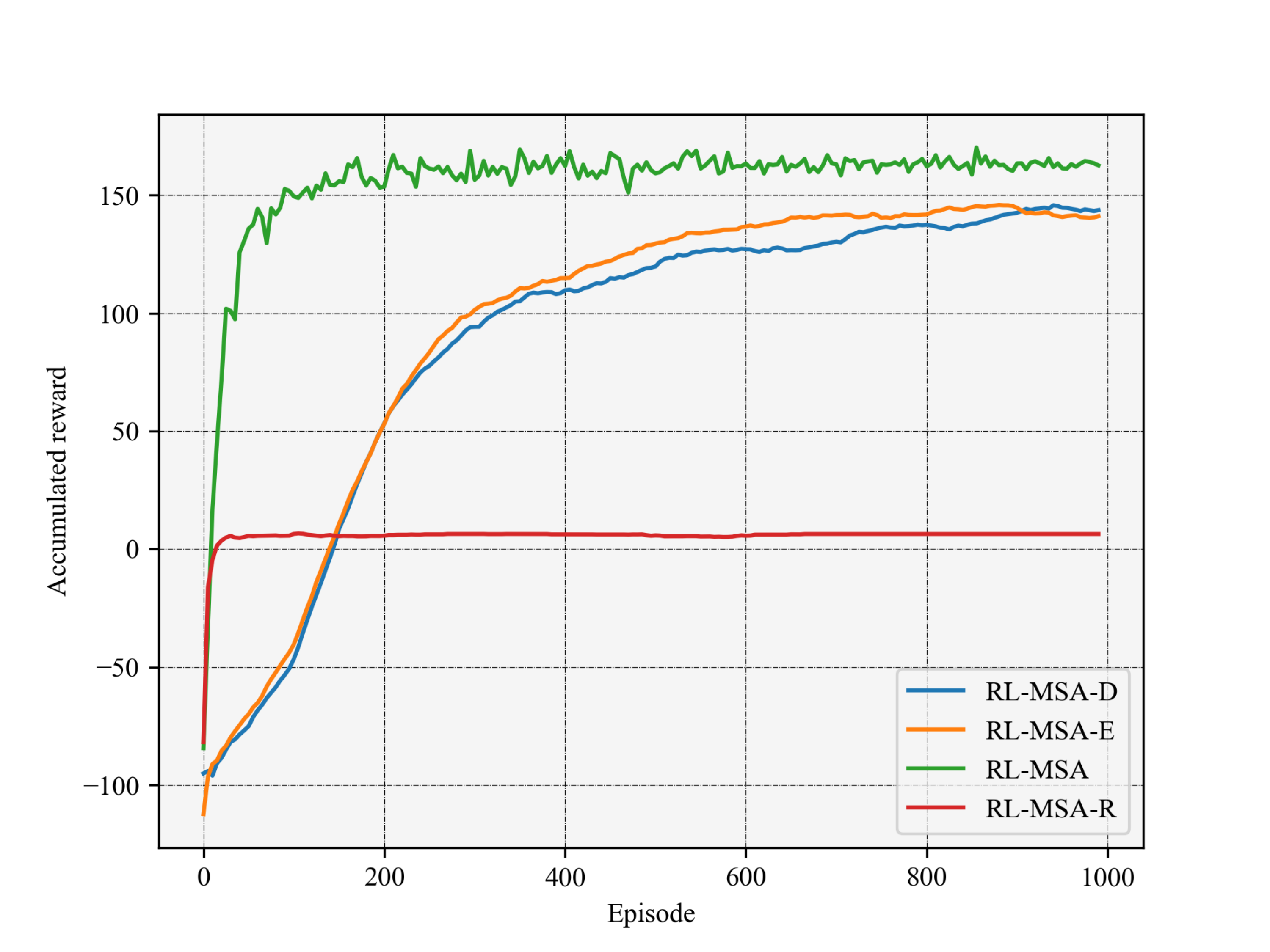}
	\centering
	\caption{Comparison of RL-MSA, RL-MSA-D, RL-MSA-E and RL-MSA-R}
\end{figure}
	
\begin{table}[H]
	\centering
	\caption{Comparison of RL-MSA, RL-MSA-D, RL-MSA-E and RL-MSA-R}
	\begin{tabular}{llll}
		\toprule[1pt]
		Approaches       & $ N_{u} $     & $ T_{d} $     & $ N_{d} $  \\ \toprule[0.7pt]   
		RL-MSA           & 46            & 0             & 0        \\
		RL-MSA-D         & 46            & 33.0          & 0        \\ 
		RL-MSA-E         & 48            & 254.0         & 0        \\
		RL-MSA-R         & 50            & 515.0         & 0        \\
		\toprule[1pt]
	\end{tabular}
\end{table}
	
\subsubsection{bus priority screening mechanism}
	
To validate the effectiveness of the bus priority screening mechanism, this paper conducts experiments on Real-1. The convergence and the performance of the model are compared between the approach using the bus priority screening mechanism (RL-MSA) and the approach without using the bus priority screening mechanism (RL-MSA-F). In RL-MSA-F, since the bus priority screening mechanism is not used, the number of buses in both the state space and action space will increase. The convergence curves of the two approaches are shown in Figure 9 (horizontal axis is the episode, and vertical axis is the accumulated reward). The three objectives achieved by the two approaches are shown in Table 7. As shown in Figure 9, firstly, the training time required by RL-MSA-F (about 954 minutes for training 400 episodes) is significantly longer than that of RL-MSA (about 21 minutes for training 400 episodes). This is because the bus priority screening mechanism reduces the size of the state space and action space, and the learning difficulty of RL agent. Secondly, RL-MSA-F still does not converge after training for 400 episodes, while RL-MSA can converge reliably after 100 episodes. In addition, the curve of RL-MSA is above the curve of RL-MSA-F, and RL-MSA finally converges to a better result than RL-MSA-F. According to the table, RL-MSA significantly outperformed RL-MSA-F in the number of buses used and the total deadhead time. In conclusion, the proposed approach achieves better results when using the bus priority screening mechanism proposed in Subsection 4.1.1.
	
\begin{figure}[H]
	\centering
	\includegraphics[height=8.2cm,width=11.3cm]{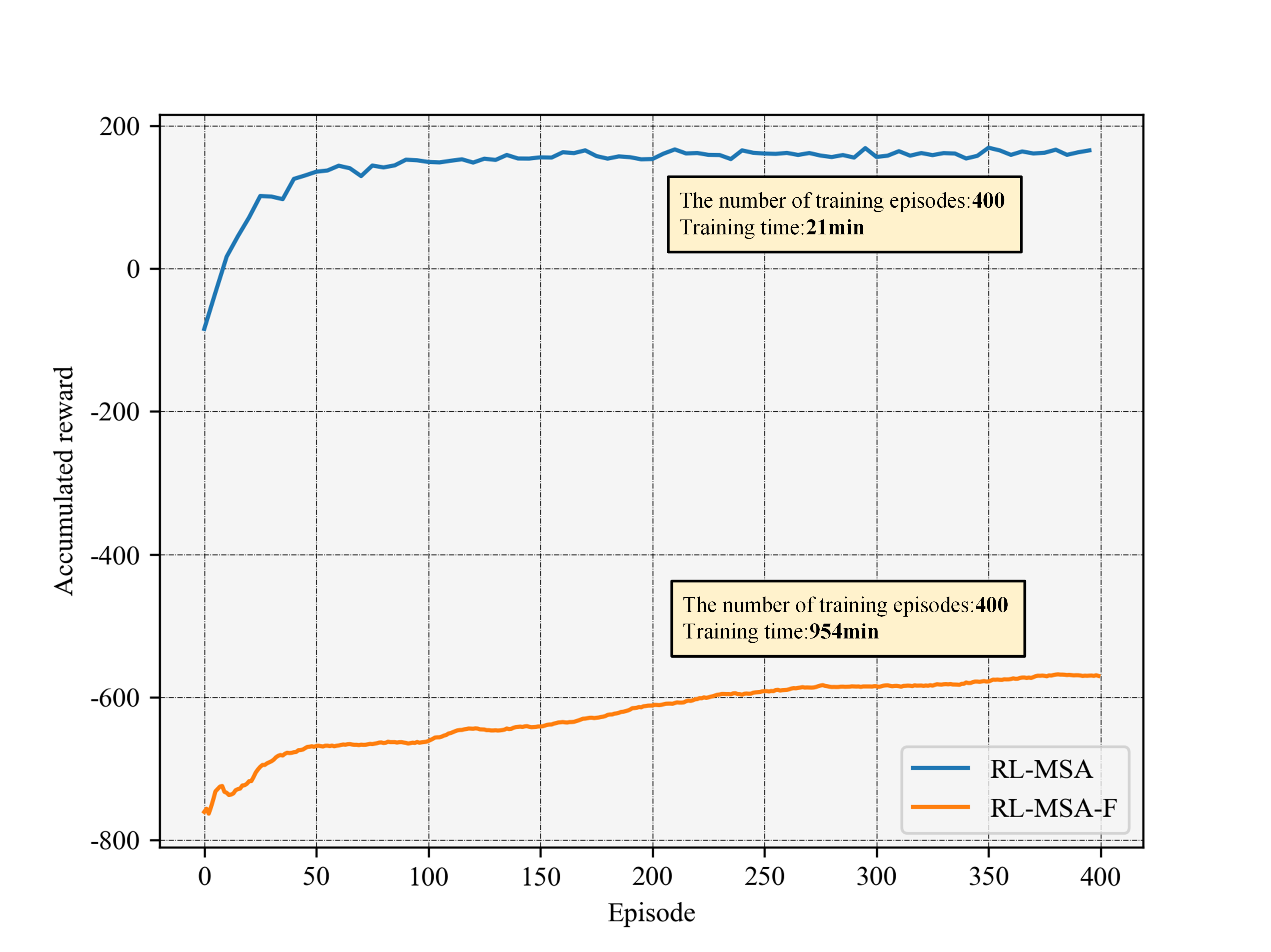}
	\centering
	\caption{Comparison of RL-MSA and RL-MSA-N}
\end{figure}
	
\begin{table}[H]
	\centering
	\caption{Comparison of RL-MSA and RL-MSA-F}
	\begin{tabular}{llll}
		\toprule[1pt]
		Approaches       & $ N_{u} $     & $ T_{d} $     & $ N_{d} $  \\ \toprule[0.7pt]   
		RL-MSA           & 46            & 0             & 0        \\
		RL-MSA-F         & 48            & 3459.0        & 0        \\
		\toprule[1pt]
	\end{tabular}
\end{table}
	
\subsubsection{Reward function}

In order to validate the effectiveness of the reward function devised in this paper, we further conduct experiments on Real-1. The performance of the model is compared between the approach using the reward function combining the final reward and the step-wise reward (RL-MSA) and the approach using only final reward (RL-MSA-C). All approaches are trained for 1000 episodes. We compare three objectives achieved by two approaches in Table 8. It can be seen from the table that only using the final reward will lead to the sparse reward problem, so the agent cannot converge fast and achieve good performance with limited training episodes. RL-MSA is better than RL-MSA-C in the number of buses used and the total deadhead time. In conclusion, RL-MSA achieves better performance when using the reward function combining the final reward and the step-wise reward.
	
\begin{table}[H]
	\centering
	\caption{Comparison of RL-MSA and RL-MSA-F}
	\begin{tabular}{llll}
		\toprule[1pt]
		Approaches       & $ N_{u} $     & $ T_{d} $     & $ N_{d} $  \\ \toprule[0.7pt]   
		RL-MSA           & 46            & 0             & 0        \\
		RL-MSA-C         & 62            & 1158.0        & 0        \\
		\toprule[1pt]
	\end{tabular}
\end{table}

\section{Conclusion}
	
Multiple Line Bus Scheduling Problem (MLBSP) is vital to save operational cost of bus company and ensure service quality for passengers. MLBSP is modeled as Markov Decision Process (MDP) for the first time. A Reinforcement Learning-based Multi-line bus Scheduling Approach (RL-MSA) is proposed and can be used at the offline and online phases. Each departure time in the bus timetable is a decision point and a RL agent makes two types of decisions (i.e., bus selection and deadhead) at each departure time according to real-time information.
	
In RL-MSA, Proximal Policy Optimization (PPO) is employed as the agent. State features include the features of buses, control points and bus lines, where the bus features are constructed based on the bus priority screening mechanism. A reward function combining the final reward and the step-wise reward is devised. At the offline phase, deadhead decision is integrated into bus selection decision. At the online phase, two types of decisions need to be made independently and deadhead decision is made through a time window mechanism based on the model trained at the offline phase. 
	
At the offline phase, experiments on 2 real-world and 8 artificial instances show that RL-MSA is superior to Adaptive Large Neighborhood Search (ALNS) in terms of the total number of buses used and total deadhead time. At the online phase, experiments show that RL-MSA can schedule buses in an online manner and cover all departure times in a timetable without increasing the number of buses used when uncertain events occur.
	
\section*{Acknowledgment}
This work was supported by National Natural Science Foundation of China under Grant 61873040.

\bibliography{main}
	
\end{document}